\documentclass[10pt,twocolumn,letterpaper]{article}

\usepackage[pagenumbers]{cvpr} 

\usepackage{amsmath}    
\usepackage{amssymb}    
\usepackage{amsfonts}   
\usepackage{mathtools}  
\usepackage{bbm}        
\usepackage{graphicx}   
\usepackage{xcolor}     
\usepackage{adjustbox}
\usepackage{booktabs, multirow, threeparttable}
\usepackage{siunitx}
\usepackage{adjustbox}

\usepackage{algorithm}      
\usepackage{algorithmic}    

\usepackage{pgfplots}
\usepackage{pgfplotstable}
\usepackage{booktabs}
\usepackage{subcaption}
\pgfplotsset{compat=1.18}
\usetikzlibrary{patterns,fit,backgrounds}
\usepgfplotslibrary{groupplots,fillbetween,colormaps}
\definecolor{mytarget}{RGB}{214,39,40}
\pgfplotsset{
  mygrid/.style={grid=both,minor tick num=1,major grid style={opacity=0.25},minor grid style={opacity=0.15}},
  myaxis/.style={font=\small, ticklabel style={font=\small}, label style={font=\small}, legend style={font=\small}},
  mylegend/.style={draw=none, fill=none, legend cell align=left, /tikz/every even column/.append style={column sep=0.3em}},
  myline/.style={line width=1.1pt, mark size=2.3pt},
  mybar/.style={ybar, bar width=10pt, line width=0.3pt},
  myerr/.style={error bars/.cd, y dir=both, y explicit, error bar style={line width=0.7pt}},
  myshade/.style={fill opacity=0.10, draw opacity=0},
  mytarget/.style={dashed, line width=0.9pt},
  cycle list name=color list,
}
\pgfplotscreateplotcyclelist{color list}{
  {black, mark=*},
  {blue!70!black, mark=square*},
  {red!70!black, mark=triangle*},
  {teal!70!black, mark=diamond*},
  {violet!80!black, mark=o},
  {orange!85!black, mark=star},
}

\usepackage{tikz}
\usetikzlibrary{arrows.meta,positioning,shapes.multipart,shadows.blur,fit,calc,backgrounds,decorations.pathreplacing}
\usepackage{fontawesome5} 
\usepackage[table]{xcolor}
\usepackage{array}      
\usepackage{siunitx}    

\definecolor{ourcol}{RGB}{220,235,255} 
\definecolor{popBlue}{HTML}{2F6BFF}
\definecolor{popTeal}{HTML}{1F9E89}
\definecolor{popOrange}{HTML}{F39C12}
\definecolor{popViolet}{HTML}{7E57C2}
\definecolor{popGray}{HTML}{6C7A89}
\definecolor{popRed}{HTML}{D35454}
\definecolor{paperBG}{HTML}{FAFBFF}

\tikzset{
  >={Stealth[length=2.2mm]},
  box/.style={draw, rounded corners=6pt, fill=white, very thick, align=center, font=\small, inner xsep=8pt, inner ysep=8pt, blur shadow},
  mini/.style={draw, rounded corners=4pt, fill=white, thick, align=center, font=\scriptsize, inner xsep=6pt, inner ysep=4pt, blur shadow},
  tag/.style={draw, rounded corners=3pt, fill=paperBG, align=center, font=\scriptsize, inner xsep=4pt, inner ysep=2pt},
  note/.style={font=\scriptsize, align=left, text=popGray},
  flow/.style={very thick, draw=popGray},
  flowBlue/.style={very thick, draw=popBlue},
  flowTeal/.style={very thick, draw=popTeal},
  legendBox/.style={draw, rounded corners=6pt, fill=white, thick, align=left, font=\scriptsize, inner xsep=8pt, inner ysep=6pt},
  badge/.style={circle, draw=popBlue, fill=popBlue!8, inner sep=1.8pt, line width=0.6pt, font=\scriptsize},
  cert/.style={draw=popBlue, very thick, rounded corners=5pt, fill=popBlue!6, inner xsep=4pt, inner ysep=2pt, font=\scriptsize},
  budget/.style={draw=popOrange, thick, rounded corners=5pt, fill=popOrange!10, inner xsep=4pt, inner ysep=2pt, font=\scriptsize},
  answerSet/.style={draw=popTeal, very thick, rounded corners=6pt, fill=popTeal!6, inner xsep=6pt, inner ysep=4pt, font=\small\bfseries},
}

\usepackage{caption}
\usepackage{subcaption}     

\definecolor{cvprblue}{rgb}{0.21,0.49,0.74}
\usepackage[pagebackref,breaklinks,colorlinks,allcolors=cvprblue]{hyperref}
\def\paperID{26934} 
\def\confName{CVPR}
\def\confYear{2026}

\title{Proof-of-Perception: Certified Tool-Using Multimodal Reasoning with Compositional Conformal Guarantees}

\author{Arya Fayyazi\\
University of Southern California\\
Los Angeles, CA\\
{\tt\small afayyazi@usc.edu}
\and
Haleh Akrami\\
Nuro\\
Mountain View, CA\\
{\tt\small hakrami@nuro.ai}
}

\begin{document}
\maketitle
\begin{abstract}
We present \textbf{Proof-of-Perception (PoP)}, a tool-using framework that casts multimodal reasoning as an executable graph with explicit reliability guarantees. Each perception or logic node outputs a conformal set $\Gamma^{(t)}_\delta(x)$, yielding calibrated, stepwise uncertainty; a lightweight controller uses these certificates to allocate compute under a budget—expanding with extra tool calls only when needed and stopping early otherwise. This grounds answers in verifiable evidence, reduces error compounding and hallucinations, and enables principled accuracy–compute trade-offs. Across document, chart, and multi-image QA benchmarks, PoP improves performance and reliability over strong chain-of-thought, ReAct-style, and program-of-thought baselines while using computation more efficiently. Code is available at \href{https://github.com/AryaFayyazi/PoP}{https://github.com/AryaFayyazi/PoP}.
\end{abstract}    
\section{Introduction}

\begin{figure*}[t]
\centering
\caption{Overview of the \textbf{Proof-of-Perception (PoP)} workflow. 
Each perception or reasoning operation is represented as a node in a directed acyclic graph (DAG). 
Each node is equipped with a conformal prediction head that provides calibrated uncertainty sets $\Gamma^{(t)}_\delta(x)$, 
and a controller adaptively allocates computation based on these certificates, producing reliable and explainable outputs.}
\label{fig:pop_workflow}

\scalebox{0.65}{
\begin{tikzpicture}[font=\sffamily, 
    node distance=2.5cm and 2.0cm, 
    >=Latex, 
    align=center,
    line width=0.7pt,
    every node/.style={align=center},
    box/.style={draw, rounded corners=8pt, fill=white, blur shadow={shadow blur steps=5}, 
                minimum height=1cm, minimum width=2.2cm, text width=2.5cm, font=\small},
    circlebox/.style={draw, circle, fill=white, blur shadow={shadow blur steps=5}, 
                      minimum size=1.0cm, inner sep=2pt},
    arrowstyle/.style={-{Latex[length=3mm, width=2mm]}, line width=0.6pt},
    labelbox/.style={font=\scriptsize\bfseries, fill=black!5, rounded corners=4pt, inner sep=2pt},
    legendbox/.style={draw, fill=white, rounded corners=4pt, font=\footnotesize, inner sep=4pt}
]

\node[box, fill=blue!10] (input) {\faImage~Images + \faKeyboard~Text Prompt};
\node[below=0.1cm of input, font=\scriptsize, text width=3.5cm, align=center] {Multimodal input \\ to MLLM system};

\node[box, fill=cyan!10, right=of input, xshift=0.8cm] (ocr) {\faEye~OCR Node \\ $\Gamma^{(\text{OCR})}_\delta(x)$};
\node[box, fill=cyan!10, below=of ocr] (det) {\faVectorSquare~Detection Node \\ $\Gamma^{(\text{Det})}_\delta(x)$};
\node[box, fill=cyan!10, above=of ocr] (chart) {\faChartBar~Chart Parsing Node \\ $\Gamma^{(\text{Chart})}_\delta(x)$};

\node[box, fill=yellow!20, right=of ocr, xshift=0.8cm] (logic) {\faProjectDiagram~Logic Fusion Node \\ $\Gamma^{(\text{Logic})}_\delta(x)$};

\node[box, fill=red!10, right=of logic, xshift=0.8cm, text width=2.7cm] (controller) {\faBrain~Adaptive Controller \\ Budget-aware Decision};

\node[box, fill=green!10, right=of controller, xshift=0.8cm, text width=2.7cm] (output) {\faCheckCircle~Certified Answer \\ with Evidence Trace};

\draw[arrowstyle] (input) -- (chart);
\draw[arrowstyle] (input) -- (ocr);
\draw[arrowstyle] (input) -- (det);
\draw[arrowstyle] (chart) -- (logic);
\draw[arrowstyle] (ocr) -- (logic);
\draw[arrowstyle] (det) -- (logic);
\draw[arrowstyle] (logic) -- (controller);
\draw[arrowstyle] (controller) -- (output);

\draw[arrowstyle, dashed, red!70!black, bend right=15] (controller.north) to node[above, font=\scriptsize, text width=1.5cm, align=center] {Retry / Expand \\ if uncertain} (chart.east);
\draw[arrowstyle, dashed, red!70!black, bend left=20] (controller) to[bend right=20] (ocr);
\draw[arrowstyle, dashed, red!70!black, bend left=15] (controller.south) to node[below, font=\scriptsize, text width=1.7cm, align=center] {Re-evaluate \\ low-confidence nodes} (det.east);

\node[legendbox, below=3.8cm of ocr, anchor=north west, text width=12.5cm] (legend) {
\textcolor{cyan!80!black}{\faEye, \faVectorSquare, \faChartBar} denote perception nodes producing calibrated sets $\Gamma^{(t)}_\delta(x)$; 
\textcolor{yellow!80!black}{\faProjectDiagram} indicates reasoning fusion nodes aggregating perception evidence; 
\textcolor{red!70!black}{\faBrain} is the adaptive controller that adjusts computation based on coverage and budget; 
and \textcolor{green!70!black}{\faCheckCircle} is the final certified output with verifiable evidence trace.
Dashed red arrows indicate feedback loops for uncertainty-driven retries or expansions.
};

\end{tikzpicture}
} 

\end{figure*}
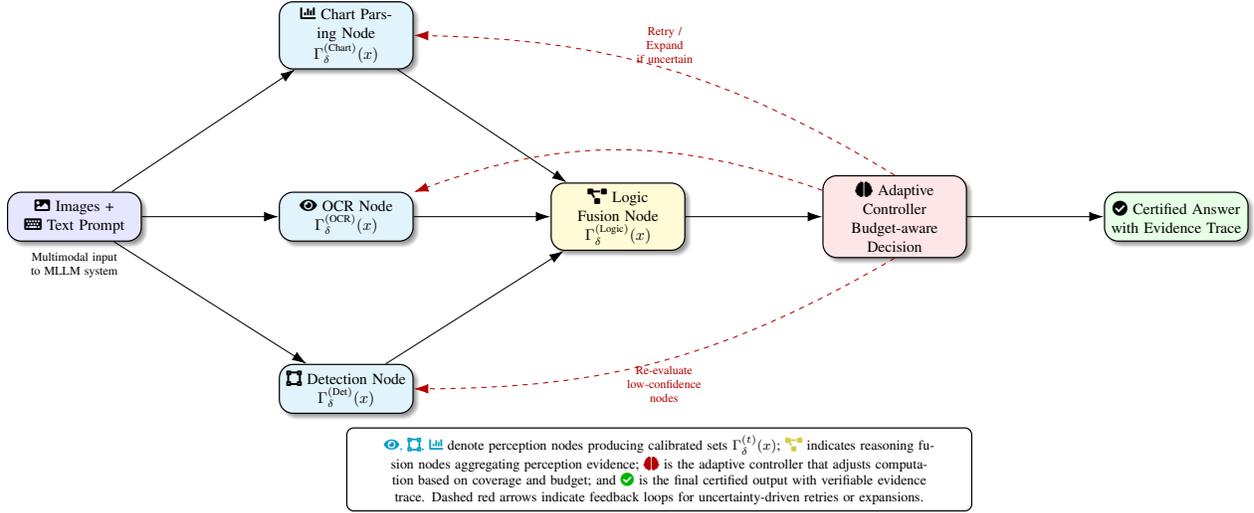

Multimodal large language models (MLLMs) have pushed open-ended vision--language tasks forward by coupling visual encoders with autoregressive language decoders \cite{alayrac2022flamingo,liu2024llava,wang2024qwen2vl}. Yet in document understanding, chart reasoning, and multi-image QA, a single forward pass still entangles fine-grained perception (OCR, detection, chart parsing) with symbolic reasoning, producing brittle, single-valued intermediates and confident but unsupported answers \cite{ji2023hallucinations,guo2017calibration}. Tool-use and structured prompting partially address this: multimodal chain-of-thought elicits textual rationales \cite{zhu2023mcot,liu2023mcotsimple}, ReAct-style agents interleave thoughts and tool calls \cite{yao2022react,yang2023mmreact}, and program-of-thought systems execute DSL programs with perception primitives \cite{nie2023programvqa,kim2024chartpot}. However, these approaches typically (i) commit to single guesses at intermediate steps, (ii) govern compute via heuristics, and (iii) calibrate, if at all, only the final answer—leaving stepwise reliability and evidence grounding unaddressed.

We present \textbf{Proof-of-Perception (PoP)}, which casts multimodal reasoning as execution of a directed acyclic graph whose nodes are perception or logic operations equipped with conformal certificates. For node type $t$ with input $x$ and candidate $z$, a learned nonconformity $s^{(t)}(x,z)$ and a split-conformal threshold $\tau^{(t)}_\delta$ define the \emph{set-valued} output $\Gamma^{(t)}_\delta(x)=\{z:\,s^{(t)}(x,z)\le\tau^{(t)}_\delta\}$, yielding marginal coverage $1-\delta$ under exchangeability \cite{vovk2005algorithmic,angelopoulos2021gentle}. A lightweight controller observes node-wise sets and a budget to decide when to accept, retry at higher fidelity, or expand with additional tool calls, turning uncertainty into a compute policy rather than a passive score. This design keeps multiple calibrated candidates until evidence resolves ambiguity, grounds answers in verifiable perceptual traces, and enables principled accuracy--compute trade-offs. In experiments on DocVQA, TextVQA, InfographicVQA, ChartQA, and MultiDoc2Dial, PoP improves performance and reliability over strong chain-of-thought, ReAct-style, and program-of-thought baselines under matched backbones and tools, while using computation more efficiently.

\section{Background and Related Work}
\label{sec:background-related}

\subsection{Multimodal Reasoning and CP}
\label{sec:mllm-cp-brief}
Modern multimodal large language models (MLLMs) couple frozen or trainable visual encoders with autoregressive decoders to model
$p_\theta(y_{1:T}\mid I_{1:M},q)$, enabling open-ended responses grounded in images and text \cite{li2023blip2,chen2023palix,peng2023kosmos2}.
Despite strong progress, directly decoding this distribution entangles \emph{fine-grained perception} (e.g., OCR, layout parsing, chart value extraction) with \emph{symbolic reasoning} (aggregation, arithmetic), which leads to brittle cascades: early perceptual slips force later steps to rationalize errors, producing confident but unsupported answers in document and chart understanding tasks \cite{kim2022donut,lee2023pix2struct,masry2022chartqa,mathew2022infographicvqa}.
Two prevalent remedies structure inference differently.
First, multimodal chain-of-thought elicits intermediate textual rationales; however, these remain free-form tokens with no guarantee they correspond to the visual evidence.
Second, tool/program-based agents let MLLMs call OCR, detection, or chart parsers (or emit a small DSL program executed by an interpreter), which improves modularity but typically commits to single-valued intermediates and governs compute with ad hoc rules (fixed retry counts, uncalibrated thresholds).
Across both families, intermediate uncertainty is rarely quantified, so decisions to expand, retry, or stop cannot be tied to principled reliability.

Conformal prediction (CP) offers distribution-free, finite-sample guarantees by turning point predictions into calibrated \emph{sets}.
Given a nonconformity function $s:\mathcal{X}\times\mathcal{Y}\to\mathbb{R}_{\ge0}$ and calibration scores $\alpha_j=s(x_j,y_j)$, split-conformal inference selects the order statistic $\tau_\delta=\alpha_{(k)}$, $k=\lceil(n+1)(1-\delta)\rceil$, and returns
$\Gamma_\delta(x)=\{y:\ s(x,y)\le\tau_\delta\}$,
which satisfies $\mathbb{P}(Y\in\Gamma_\delta(X))\ge 1-\delta$ under exchangeability \cite{lei2018distributionfree,bates2023rcps, fayyazi2025facterfairnessawareconformalthresholding}.
While recent work has applied CP to final predictions in classification and detection~\cite{Fayyazi_2026_WACV}, most applications treat the model as monolithic ($x\mapsto y$), leaving multi-step tool outputs uncalibrated and disconnected from compute policies.
Our approach brings CP \emph{inside} multimodal reasoning: each node in a directed acyclic graph (OCR, detection, chart parsing, logic fusion) emits a set-valued output $\Gamma^{(t)}_\delta(x_v)$ with coverage guarantees, and a lightweight controller uses these certificates and a budget to decide whether to accept, retry at higher fidelity, or expand with additional tool calls.
This integrates reliability with computation: uncertainty actively routes effort to ambiguous subproblems, while confident segments terminate early, yielding evidence-grounded answers and controllable accuracy--compute trade-offs.

\subsection{Limitations and Positioning}
\label{sec:limits-positioning}
Existing multimodal pipelines face three persistent issues in complex visual reasoning: (i) \textit{single-valued intermediates} (one OCR string, one box, one chart value) that silently propagate errors; (ii) \textit{heuristic compute control} (fixed depths, template programs, uncalibrated thresholds) that cannot tune accuracy--compute trade-offs; and (iii) a lack of \textit{compositional reliability}, since any calibration is typically applied only to final answers, not to the sequence of perception and logic steps that produce them.
\textbf{Proof-of-Perception (PoP)} addresses these gaps while remaining compatible with standard MLLMs and tools.
For each node type $t\!\in\!\{\mathrm{OCR,Det,Chart,Logic}\}$ with input state $x_v$ and candidate $z$, PoP learns a nonconformity function $s^{(t)}(x_v,z)$ and, via split-conformal calibration, outputs
$\Gamma^{(t)}_\delta(x_v)=\{z:\ s^{(t)}(x_v,z)\le\tau^{(t)}_\delta\}$,
so that every intermediate operation is certified rather than point-valued.
A controller observes node-wise set geometry (e.g., size, dispersion) and a global budget to accept, retry at higher fidelity (e.g., higher-resolution crop, alternate tool parameters), or expand the reasoning graph with additional tool calls.
By tying computation to certified uncertainty, PoP reduces error compounding and hallucinations, grounds answers in verifiable perceptual evidence, and exposes a knob to trade accuracy for cost across document, chart, and multi-image QA settings.

\section{Methodology}
\label{sec:method}

In this section, we first define the problem and notation, then describe the reasoning graph representation, the node-level conformal prediction mechanism, the adaptive controller, the self-play counterexample miner, and finally the training and inference algorithms.

\subsection{Problem Setup and Notation}

We consider a generic multimodal reasoning task with inputs
\begin{align}
  x = (I_{1:M}, q),
\end{align}
where $I_{1:M} = (I_1, \dots, I_M)$ is a sequence of images (e.g., pages of a document, charts, or screenshots), and $q$ is a natural language query.
The desired output is an answer $y$ from a task-specific space $\mathcal{Y}$ (e.g., short text, numeric value, or categorical label).

We assume access to:
\begin{itemize}
  \item A base multimodal large language model (MLLM) $F_\theta$ with parameters $\theta$, which can process images and text and generate text tokens.
  \item A finite set of external perception tools
  \begin{align}
    \mathcal{T} = \{ T^{(1)}, \dots, T^{(K)} \},
  \end{align}
  where each tool $T^{(k)}$ takes as input an image (or image crop) and a text prompt, and returns a structured output in a tool-specific space $\mathcal{Z}^{(k)}$.
  Examples include OCR, object detectors, layout analyzers, chart parsers, and simple visual question answerers.
\end{itemize}

We are given a dataset
\begin{align}
  \mathcal{D} = \{ (x_i, y_i) \}_{i=1}^N
\end{align}
for training and calibration.
We partition $\mathcal{D}$ into a \emph{proper training set} $\mathcal{D}_{\text{train}}$ and a \emph{calibration set} $\mathcal{D}_{\text{cal}}$ to enable split conformal prediction.

PoP learns:
\begin{itemize}
  \item A \emph{planner} and \emph{fuser} based on $F_\theta$ that generate and execute a directed acyclic graph (DAG) of reasoning steps.
  \item Node-type-specific \emph{certificate heads} that output nonconformity scores for conformal prediction.
  \item A \emph{controller} $\pi_\phi$ that decides whether to expand the reasoning graph (e.g., add more tool calls) given current certificates and a computation budget.
\end{itemize}

\subsection{Reasoning Graph Representation}
\label{sec:graph}

We represent the reasoning process for a single input $x$ by a directed acyclic graph
\begin{align}
  G = (V, E),
\end{align}
where $V$ is a set of nodes and $E \subseteq V \times V$ is a set of directed edges.
Each node $v \in V$ corresponds to an operation that produces an intermediate random variable $Z_v$.
We consider two main node types:
\begin{itemize}
  \item \textbf{Tool nodes} $v \in V_{\text{tool}}$: nodes that call an external tool $T^{(k)} \in \mathcal{T}$.
        Each tool node is associated with:
        \begin{itemize}
          \item A tool index $k(v) \in \{1,\dots,K\}$,
          \item A region specification $r_v$ (e.g., an image index and bounding box),
          \item A textual prompt $p_v$ describing the sub-task for the tool.
        \end{itemize}
        The tool executes as
        \begin{align}
          Z_v = T^{(k(v))}(I_{1:M}, r_v, p_v) \in \mathcal{Z}^{(k(v))}.
        \end{align}
  \item \textbf{Fusion nodes} $v \in V_{\text{fuse}}$: nodes that operate purely inside the MLLM, fusing the query $q$ with intermediate results from predecessor nodes.
        Let $\text{pa}(v)$ denote the set of parent nodes of $v$.
        We model
        \begin{align}
          Z_v = f_{\theta}^{(\text{fuse})}\!\big( q, x, \{ \hat{Z}_u \}_{u \in \text{pa}(v)} \big),
        \end{align}
        where $f_{\theta}^{(\text{fuse})}$ is implemented via a forward pass of $F_\theta$ over a structured prompt that includes $q$ and textual or serialized representations of $\hat{Z}_u$.
\end{itemize}

A special \emph{answer node} $v^\star \in V_{\text{fuse}}$ produces the final answer $Z_{v^\star} \in \mathcal{Y}$.

\paragraph{Graph generation.}
Given $(I_{1:M}, q)$, the planner uses the MLLM to autoregressively generate a textual program that encodes the graph $G$.
We define a small domain-specific language (DSL) of instructions such as:
\begin{itemize}
  \item \texttt{CALL\_TOOL(k, region, prompt) $\rightarrow$ $v$}
  \item \texttt{FUSE(parents, prompt) $\rightarrow$ $v$}
  \item \texttt{RETURN(node)}
\end{itemize}
The DSL is parsed into a DAG by a deterministic interpreter.
The generated $G$ must be acyclic and respect a topological ordering.

For simplicity, we denote by $\sigma: V \rightarrow \{1,\dots,|V|\}$ a topological order of nodes.

\subsection{Node-Level Predictions and Nonconformity Scores}
\label{sec:node-nonconf}

Each node $v$ outputs a random variable $Z_v \in \mathcal{Z}_v$ from a node-specific space $\mathcal{Z}_v$.
For instance, OCR tools output strings, detectors output bounding boxes, chart parsers output numeric values, and fusion nodes output textual or categorical decisions.

For each node type $t \in \mathcal{T}_{\text{node}}$ (e.g., ``OCR-string'', ``det-box'', ``chart-value'', ``logic-text''), we define:
\begin{itemize}
  \item A base predictor
  \begin{align}
    f_{\theta}^{(t)} : \mathcal{X}_v \rightarrow \hat{\mathcal{Z}}_v,
  \end{align}
  which maps node inputs to a predicted output $\hat{z}_v$ (e.g., a probability distribution over tokens or boxes).
  Here $\mathcal{X}_v$ denotes the node-specific input state (e.g., image crop, prompt, and context).
  \item A \emph{nonconformity function}
  \begin{align}
    s^{(t)} : \mathcal{X}_v \times \mathcal{Z}_v \rightarrow \mathbb{R}_{\ge 0}
  \end{align}
  that measures how ``strange'' a candidate output $z$ is relative to $f_{\theta}^{(t)}$ and the input.
\end{itemize}
\paragraph{Examples of nonconformity.}
\begin{itemize}
  \item For string-valued outputs (e.g., OCR), we may define
  \begin{align}
    s^{(\text{ocr})}(x_v, z) = 1 - P_{\theta}(z \mid x_v),
  \end{align}
  where $P_{\theta}(z \mid x_v)$ is the model probability of string $z$.
  Alternatively, we can use edit distance to the MAP prediction.
  \item For bounding boxes, we can define
  \begin{align}
    s^{(\text{box})}(x_v, z) = 1 - \text{IoU}\big( z, \hat{z}_v^{\text{MAP}} \big),
  \end{align}
  where $\hat{z}_v^{\text{MAP}}$ is the highest-scoring box.
  \item For scalar numeric outputs, we can define a residual-based score
  \begin{align}
    s^{(\text{num})}(x_v, z) = \big| z - \mu_\theta(x_v) \big|
  \end{align}
  for a predicted mean $\mu_\theta(x_v)$.
\end{itemize}

During training, when ground-truth intermediate labels are available (e.g., OCR transcripts or chart values), we can compute:
\begin{align}
  s^{(t)}(x_v, z_v^{\text{true}}) \quad \text{for each labeled node $v$ of type $t$.}
\end{align}
When intermediate labels are not available, we obtain pseudo-labels via teacher models or task-consistent heuristics (e.g., parsing gold answers back into node-level values).

\subsection{Split CP for Node Certificates}
\label{sec:cp}

We now describe how to turn node-level predictors into calibrated \emph{set-valued} predictors with provable marginal coverage guarantees, using split conformal prediction.
\paragraph{Calibration data.}
We collect a set of node-level calibration examples for each node type $t$:
\begin{align}
  \mathcal{C}^{(t)} = \{ (x^{(t)}_j, z^{(t)}_j) \}_{j=1}^{n_t},
\end{align}
where $(x^{(t)}_j, z^{(t)}_j)$ are node inputs and their corresponding (true or pseudo-true) outputs, extracted from the calibration split $\mathcal{D}_{\text{cal}}$.
For each $(x^{(t)}_j, z^{(t)}_j)$, we compute the nonconformity score
\begin{align}
  \alpha^{(t)}_j = s^{(t)}\big(x^{(t)}_j, z^{(t)}_j\big).
\end{align}

Let $\alpha^{(t)}_{(1)} \le \alpha^{(t)}_{(2)} \le \dots \le \alpha^{(t)}_{(n_t)}$ denote the sorted scores.
Given a desired marginal coverage level $1 - \delta$ with $\delta \in (0,1)$, split conformal prediction chooses the threshold
\begin{equation}
  \label{eq:cp-threshold}
  \tau^{(t)}_{\delta}
  = \alpha^{(t)}_{(k)}, 
  \quad
  k = \Big\lceil (n_t + 1)(1 - \delta) \Big\rceil.
\end{equation}
\paragraph{Set-valued prediction.}
Given a new node input $x_v$ of type $t$, we define the conformal set
\begin{equation}
  \label{eq:cp-set}
  \Gamma^{(t)}_{\delta}(x_v)
  = \big\{
      z \in \mathcal{Z}_v 
      : s^{(t)}(x_v, z) \le \tau^{(t)}_{\delta}
    \big\}.
\end{equation}
Under the standard assumption that the calibration and test node inputs are exchangeable and that nonconformity scores are computed in the same way, the split conformal construction guarantees
\begin{align}
  \mathbb{P}\Big( z_v^{\text{true}} \in \Gamma^{(t)}_{\delta}(x_v) \Big)
  \ge 1 - \delta,
\end{align}
where the probability is over the randomness of training/calibration data and the test node input.
\paragraph{Practical instantiation.}
In practice, many $\mathcal{Z}_v$ are discrete but large (e.g., strings).
We approximate \eqref{eq:cp-set} with a finite candidate set $\mathcal{Z}^{\text{cand}}_v$ generated from the base model (e.g., via beam search or sampling).
We then include $z \in \mathcal{Z}^{\text{cand}}_v$ in $\Gamma^{(t)}_{\delta}(x_v)$ if $s^{(t)}(x_v, z) \le \tau^{(t)}_{\delta}$.
For ease of computation, we can enforce a maximum allowed set size $K_{\max}$ by truncating to the $K_{\max}$ least nonconforming candidates.
\paragraph{Certificate head.}
We implement each nonconformity function $s^{(t)}$ with a small \emph{certificate head} on top of the node representation:
\begin{align}
  s^{(t)}(x_v, z) 
  = g^{(t)}_\psi \big( \phi^{(t)}_\theta(x_v, z) \big),
\end{align}
where $\phi^{(t)}_\theta$ is a feature extractor (e.g., the hidden state of the MLLM or a vision backbone), $g^{(t)}_\psi$ is a shallow MLP, and $\psi$ are certificate-head parameters.
This allows us to jointly train the base predictor and the certificate head.

\subsection{Adaptive Controller for Computation Allocation}
\label{sec:controller}

Running many tool calls and re-parsing images can be expensive.
PoP includes a controller that decides, per node, whether to accept the current conformal set or to \emph{expand} the reasoning graph by invoking additional operations.

For each node $v$, define the \emph{certificate state}
\begin{align}
c_v
  = \big(
    &\tau^{(t(v))}_{\delta},\;
      \widehat{\mathbbm{1}}\!\left[z_v^{\text{true}} \in 
      \Gamma^{(t(v))}_{\delta}(x_v)\right], \nonumber\\
    &|\Gamma^{(t(v))}_{\delta}(x_v)|,\;
      \text{type}(v)
    \big).
\end{align}

where $t(v)$ is the node type.
At test time, the ground-truth indicator is not known; it is only used during training to define learning signals (see \S\ref{sec:training}).
We also maintain a \emph{global budget state} $b$ recording the cumulative cost of tool calls, image crops, and model passes.

The controller $\pi_\phi$ is a policy parameterized by $\phi$:
\begin{align}
  a_v = \pi_\phi\big( c_v, b, \text{context}(v) \big),
\end{align}
where $\text{context}(v)$ aggregates local information, such as:
\begin{itemize}
  \item Encodings of the node inputs,
  \item Loose summaries of upstream nodes (e.g., a pooled embedding over $\{\hat{Z}_u\}_{u \in \text{pa}(v)}$),
  \item The text of $q$.
\end{itemize}

We consider a small discrete action space
\begin{align}
  \mathcal{A} = \{\texttt{ACCEPT},\; \texttt{RETRY},\; \texttt{EXPAND},\; \texttt{ABORT}\},
\end{align}
with the following semantics:
\begin{itemize}
  \item \texttt{ACCEPT}: keep the current conformal set $\Gamma^{(t(v))}_{\delta}(x_v)$ as the node output.
  \item \texttt{RETRY}: re-run the same node with a higher-quality configuration (e.g., higher resolution crop or alternative tool parameters).
  \item \texttt{EXPAND}: create new child nodes that refine the current node (e.g., extra OCR calls on subregions, or parallel tools).
  \item \texttt{ABORT}: early stop if the controller judges the query unanswerable under the budget, returning a special ``no answer'' token.
\end{itemize}

The controller is trained to trade off task performance and computation cost.
\subsection{Self-Play Counterexample Mining}
\label{sec:selfplay}
To harden robustness and enrich calibration with difficult cases, PoP uses a self-play loop between a trainable \emph{student} $(\theta,\psi,\phi)$ and a periodically frozen \emph{adversary} $(\tilde{\theta},\tilde{\psi},\tilde{\phi})$ cloned from the student. For each example $(x,y)$, the adversary proposes and executes a reasoning graph $\tilde{G}$, then generates perturbed inputs and intermediate states $(x',\{z'_v\})$ via controlled visual/layout/tool shifts (e.g., crops, mild affine changes, distractors, injected OCR noise), and filters for \emph{counterexamples} where $\tilde{y}\neq y$ or node nonconformity is large. The student is trained on these cases to recover the correct answer, to produce node-wise conformal sets $\Gamma^{(t)}_\delta(x_v')$ that achieve the target coverage, and to learn controller actions that maintain coverage without overspending compute. The selected node-level pairs $(x_v',z_v')$ are appended to calibration pools $\mathcal{C}^{(t)}$ so that thresholds $\tau^{(t)}_\delta$ reflect realistic failure modes, yielding certificates and policies that remain reliable under distributional perturbations.

\subsection{Training Objective}
\label{sec:training}

The overall training objective for PoP combines a task loss, a planning loss, a certificate loss, a controller loss, and a budget regularizer.
\paragraph{Task loss.}
For each training example $(x, y)$, we execute the student model with a fixed or softly supervised graph and obtain the final prediction $\hat{y}$.
We define a task loss
\begin{align}
  \mathcal{L}_{\text{task}} = \mathbb{E}_{(x,y) \sim \mathcal{D}_{\text{train}}}
  \big[ \ell_{\text{task}}(\hat{y}, y) \big],
\end{align}
where $\ell_{\text{task}}$ is, for instance, cross-entropy for classification or a smooth-$L_1$ loss for regression.
\paragraph{Planning loss.}
When ground-truth or teacher-provided programs (graphs) are available, we supervise the planner with a sequence-level loss.
Let $p_\theta(\pi \mid x)$ denote the probability of emitting program $\pi$ (i.e., the textual DSL for $G$).
Given a reference program $\pi^\star$, we define
\begin{align}
  \mathcal{L}_{\text{plan}} = \mathbb{E}_{(x,\pi^\star)} \big[ -\log p_\theta(\pi^\star \mid x) \big].
\end{align}
When reference programs are not available, we can define a reinforcement-style reward for graphs that achieve high accuracy with low cost, and optimize $p_\theta$ with policy gradients; for brevity we do not expand that formulation here.
\paragraph{Certificate loss.}
For each node type $t$ and labeled node example $(x_v, z_v)$, we want the learned nonconformity scores $s^{(t)}(x_v,z_v)$ to align with the empirical quantile thresholds $\tau^{(t)}_\delta$.
A simple surrogate is a margin-based loss:
\begin{align}
  \ell_{\text{cert}}^{(t)}(x_v, z_v)
   =
   \max\big\{ 0,\; s^{(t)}(x_v,z_v) - \tau^{(t)}_\delta - \epsilon \big\},
\end{align}
where $\epsilon \ge 0$ is a small slack parameter that encourages scores of true outputs to fall below the learned threshold.
We aggregate over all node types and examples:
\begin{align}
  \mathcal{L}_{\text{cert}}
  =
  \sum_{t} \lambda_t
  \, \mathbb{E}_{(x_v, z_v) \sim \mathcal{C}^{(t)}} 
  \big[ \ell_{\text{cert}}^{(t)}(x_v, z_v) \big],
\end{align}
with non-negative weights $\lambda_t$.
\paragraph{Controller loss.}
We define a per-example cost
\begin{align}
  C(x) = C_{\text{err}}(x) + \beta \, C_{\text{comp}}(x),
\end{align}
where:
\begin{itemize}
  \item $C_{\text{err}}(x)$ penalizes incorrect final answers or coverage violations (e.g., if $y \notin \Gamma^{(\text{answer})}_\delta$ at the answer node).
  \item $C_{\text{comp}}(x)$ measures computation cost, for example
  \begin{align}
    C_{\text{comp}}(x)
    = \sum_{v \in V_{\text{tool}}} c_{\text{tool}}(v)
      + \sum_{v \in V_{\text{fuse}}} c_{\text{fuse}}(v),
  \end{align}
  where $c_{\text{tool}}(v)$, $c_{\text{fuse}}(v)$ are pre-defined costs per tool call or fusion pass.
\end{itemize}
The hyperparameter $\beta > 0$ trades off accuracy and cost.

We optimize the controller parameters $\phi$ to minimize the expected cost:
\begin{align}
  \mathcal{L}_{\text{ctrl}} = \mathbb{E}_{(x,y) \sim \mathcal{D}_{\text{train}}}
  \big[ C(x) \big].
\end{align}
Since decisions $a_v$ are discrete, we use a policy-gradient estimator.
Let $R(x) = -C(x)$ be the reward.
The gradient with respect to $\phi$ is approximated as
\begin{align}
\nabla_\phi \mathcal{L}_{\text{ctrl}}
  &\approx 
  - \mathbb{E}\Bigg[
    \sum_{v \in V}
      \nabla_\phi \log \pi_\phi(a_v \mid c_v, b, \text{context}(v)) \nonumber\\
  &\hspace{1.8cm}\times \big(R(x) - b_0\big)
  \Bigg].
\end{align}

where $b_0$ is a baseline (e.g., an exponential moving average of rewards).
\paragraph{Overall objective.}
The total loss is
\begin{align}
  \mathcal{L}
  = \mathcal{L}_{\text{task}}
    + \gamma_{\text{plan}} \mathcal{L}_{\text{plan}}
    + \gamma_{\text{cert}} \mathcal{L}_{\text{cert}}
    + \gamma_{\text{ctrl}} \mathcal{L}_{\text{ctrl}},
\end{align}
with non-negative weights $\gamma_{\text{plan}}, \gamma_{\text{cert}}, \gamma_{\text{ctrl}}$.

\subsection{Inference Procedure}
\label{sec:inference}

At test time, PoP operates in two main phases: (1) graph generation; (2) graph execution with node-wise conformal certificates and controller-driven expansion.
\paragraph{Phase 1: Graph generation.}
Given input $(I_{1:M}, q)$, the planner (MLLM) autoregressively generates a program $\pi$ in the DSL.
We parse $\pi$ into a DAG $G = (V, E)$ that satisfies syntactic and acyclicity constraints.
If the generated program is invalid, we fall back to a default single-pass reasoning template (e.g., direct answer with no tools).
\paragraph{Phase 2: Graph execution.}
We execute nodes in topological order $\sigma$.
For each node $v$:
\begin{enumerate}
  \item Construct the node input $x_v$ from $(I_{1:M}, q)$ and the (set-valued) outputs of its parents $\{ \Gamma^{(t(u))}_\delta(x_u) \}_{u \in \text{pa}(v)}$.
  \item Use the node-type predictor $f_\theta^{(t(v))}$ to obtain candidate outputs $\mathcal{Z}_v^{\text{cand}}$ and corresponding nonconformity scores under $s^{(t(v))}$.
  \item Compute the conformal set $\Gamma^{(t(v))}_\delta(x_v)$ using the precomputed threshold $\tau^{(t(v))}_\delta$ and the procedure in \eqref{eq:cp-set}.
  \item Query the controller $\pi_\phi$ with the certificate state $c_v$, global budget $b$, and context to decide action $a_v \in \mathcal{A}$.
  \item If $a_v = \texttt{ACCEPT}$, proceed to the next node; if $a_v \in \{\texttt{RETRY}, \texttt{EXPAND}\}$, modify the graph locally (e.g., add nodes or re-run $v$ with modified configuration); if $a_v = \texttt{ABORT}$, terminate with a special no-answer.
\end{enumerate}

After visiting all nodes, we read the final conformal set at the answer node $v^\star$:
\begin{align}
  \Gamma^{(\text{answer})}_\delta(x) 
  = \Gamma^{(t(v^\star))}_\delta(x_{v^\star})
  \subseteq \mathcal{Y}.
\end{align}
We can either:
\begin{itemize}
  \item Return the full set $\Gamma^{(\text{answer})}_\delta(x)$ as a calibrated uncertainty summary, or
  \item Select a single answer $\hat{y}$ from the set (e.g., the highest-probability element) and expose both $\hat{y}$ and $\Gamma^{(\text{answer})}_\delta(x)$ to the user.
\end{itemize}

\section{Experiments}
\label{sec:experiments}

We evaluate \emph{Proof-of-Perception (PoP)} on multimodal tasks that require \textbf{(i)} fine-grained perception (OCR, detection, layout/figure parsing), \textbf{(ii)} multi-step reasoning over intermediate evidence, and \textbf{(iii)} reliability under controlled or natural distribution shift. We adhere to the methodology in Section~\ref{sec:method}, using identical training/decoding budgets across systems unless stated. Our experiments answer four questions:

\textbf{Q1}~Does PoP improve answer quality while \emph{reducing} hallucinations? \quad
\textbf{Q2}~Do our \emph{node-wise conformal certificates} achieve the target coverage across node types and datasets? \quad
\textbf{Q3}~Can PoP \emph{allocate computation adaptively}, achieving better accuracy--compute trade-offs? \quad
\textbf{Q4}~Is PoP robust to realistic shifts (layout/fonts/clutter) via self-play counterexample mining?

\subsection{Tasks and Datasets}
\label{sec:datasets}
We evaluate where tool use and compositional perception are pivotal. \textbf{Document understanding:} \textit{DocVQA} (Task~1 \& 3) for key–value extraction and document QA on scanned forms, \textit{TextVQA} for OCR-heavy visual QA, and \textit{InfographicVQA} for dense multi-panel graphics \cite{mathew2021docvqa,singh2019textvqa,mathew2022infographicvqa}. We report Exact Match (EM), F1, and \textit{Hallucination Rate} (fraction of answers unsupported by any evidence node; Sec.~\ref{sec:hallucination-metric}). \textbf{Chart and figure reasoning:} \textit{ChartQA} (Human/Machine splits) and \textit{ChartQA-PoT} (program-of-thought annotations) evaluate numerical/semantic reasoning; metrics: EM and absolute value error \cite{masry2022chartqa,kim2024chartpot}. \textbf{Multi-image aggregation:} \textit{MultiDoc2Dial} assesses grounded QA over multiple related pages, and \textit{TextCaps} evaluates captioning with OCR grounding; we use EM (QA) or CIDEr (captioning) plus hallucination rate \cite{feng2021multidoc2dial,patel2020textcaps}.

\subsection{Models and Baselines}
\label{sec:baselines}
\textbf{PoP (ours).} The planner–fuser is a 7–9B open MLLM; tools include OCR (PaddleOCR), detection (DETR-like), a chart parser, and a layout parser \cite{du2020paddleocr,carion2020detr,shen2021layoutparser}. Node types are \texttt{ocr-string}, \texttt{det-box}, \texttt{chart-num}, and \texttt{logic-text}. Each node has a certificate head defining nonconformity $s^{(t)}$; calibration follows split CP (Sec.~\ref{sec:cp}). The controller $\pi_\phi$ chooses \{\texttt{ACCEPT}, \texttt{RETRY}, \texttt{EXPAND}, \texttt{ABORT}\} (Sec.~\ref{sec:controller}).

\textbf{Baselines.} (1) \textit{Direct-MLLM}: same backbone, no tools, single-pass decoding. (2) \textit{M-CoT}: multimodal chain-of-thought prompting (no tools). (3) \textit{MM-ReAct}: tool-using agent with heuristic retries (no calibration). (4) \textit{ProgVLM}: program-of-thought tool plans without certificates. We also include two 2025 open-source planner$\to$tools$\to$fuser agents under matched budgets.%
\footnote{All methods use the same backbone class, tokenizer, and image encoder resolution; tool implementations are shared unless the baseline is tool-free.}

\begin{figure*}[!t]
\centering
\caption{Node-wise conformal coverage vs.\ average set size across pooled datasets (target $90\%$). Bars show mean coverage; thin caps indicate $\pm1.0\%$ variation. We constrain set sizes (OCR: up to 5, boxes: up to 3), which keeps coverage near target without inflating candidate sets.}
\label{fig:coverage_bars}
\begin{tikzpicture}
\begin{axis}[
  myaxis, mygrid,
  width=0.98\linewidth, height=6.0cm,
  ymin=84, ymax=96,
  ytick={84,86,88,90,92,94,96},
  ylabel={Empirical coverage (\%)},
  xlabel={Average set size},
  xtick=data,
  xticklabels={1,2,3,4,5},
  legend style={mylegend, at={(0.5,1.04)}, anchor=south, legend columns=4},
]
\addplot+[mybar, fill=blue!22!white, draw=blue!60!black, myerr]
  coordinates {(1,87.5) +- (0,1.0) (2,90.1) +- (0,0.8) (3,91.2) +- (0,1.0) (4,91.0) +- (0,0.9) (5,90.7) +- (0,0.9)};
\addlegendentry{OCR (string)}

\addplot+[mybar, fill=teal!20!white, draw=teal!60!black, myerr]
  coordinates {(1,88.3) +- (0,0.9) (2,90.5) +- (0,0.8) (3,91.3) +- (0,0.8) (4,0) +- (0,0) (5,0) +- (0,0)};
\addlegendentry{Detection (box)}

\addplot+[mybar, fill=violet!18!white, draw=violet!70!black, myerr]
  coordinates {(1,90.2) +- (0,0.7) (2,90.5) +- (0,0.7) (3,0) +- (0,0) (4,0) +- (0,0) (5,0) +- (0,0)};
\addlegendentry{Chart (numeric)}

\addplot+[mybar, fill=orange!20!white, draw=orange!70!black, myerr]
  coordinates {(1,88.9) +- (0,0.9) (2,90.3) +- (0,0.9) (3,90.9) +- (0,1.0) (4,90.8) +- (0,1.0) (5,0) +- (0,0)};
\addlegendentry{Logic (text)}

\addplot[mytarget] coordinates {(0.5,90) (5.5,90)};
\end{axis}
\end{tikzpicture}
\end{figure*}
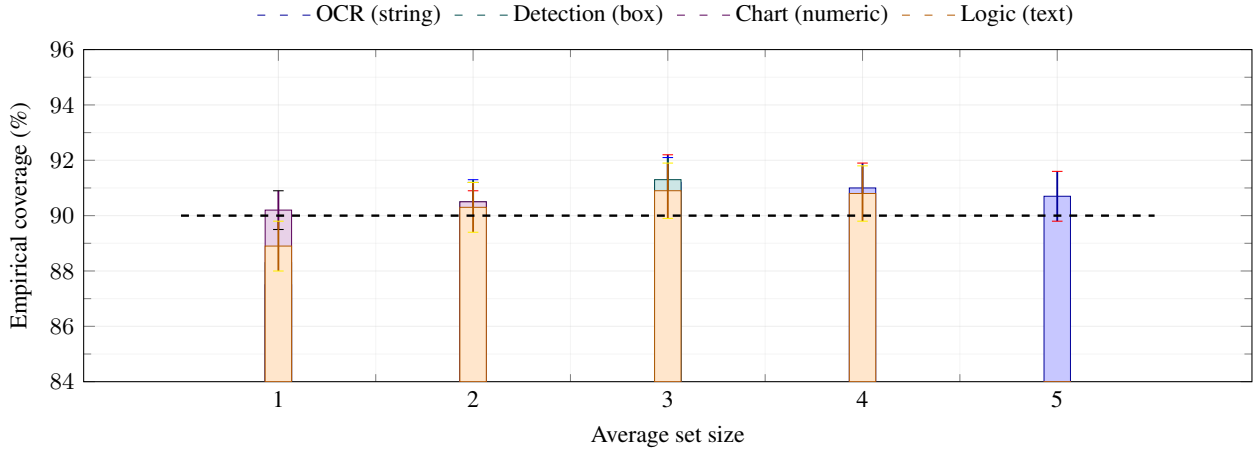
\subsection{Implementation and Training Setup}
\label{sec:setup}

\textbf{Backbone.} 7--9B class MLLM; image encoder at $896^2$; planner beam size 5; max program length 128 DSL tokens.

\textbf{Tools.} OCR with lexicon-free decoding; detector at 900 queries; chart parser with template-free axis/legend detection; layout parser trained on RVL-CDIP-style documents.

\textbf{Calibration.} Per node type $t$, we extract $n_t\!\in\![8\text{k}, 22\text{k}]$ labeled/pseudo-labeled nodes from held-out splits. Threshold $\tau^{(t)}_\delta$ uses Eq.~(1) with $\delta=0.1$ (90\% target coverage). We limit set size to $K_{\max}=5$ for \texttt{ocr-string}, $K_{\max}=3$ for \texttt{det-box}, and numeric intervals for \texttt{chart-num}.

\textbf{Controller/budget.} Each tool call costs $1$ unit, high-res retry costs $2$, and a fuse step costs $0.25$. Unless stated, the per-sample budget is $B\!=\!16$. Training uses policy gradients with reward $R=-C_{\text{err}}-\beta C_{\text{comp}}$ ($\beta=0.05$).

\textbf{Self-play counterexamples.} Every 2 epochs we refresh a frozen copy to generate perturbed instances (clutter, affine text warp, font swaps, distractor boxes, re-shuffled panels). Hard nodes join the calibration pools (Sec.~\ref{sec:selfplay}).

\subsection{Metrics}
\label{sec:metrics}

\textbf{Answer Quality.} EM / F1 (QA); Abs.\ Error (numeric).  
\textbf{Coverage (node \& answer).} Empirical coverage $\widehat{\mathrm{Cov}}=\tfrac{1}{N}\sum \mathbf{1}[z^{\star}\in \Gamma_\delta]$ vs.\ target $1-\delta=90\%$.  
\textbf{Hallucination Rate.} \label{sec:hallucination-metric}
An answer is ``hallucinated'' if it cannot be justified by any accepted node evidence along the execution trace: for text answers, no \texttt{ocr-string} or \texttt{logic-text} node entails the claimed span; for numeric answers, no \texttt{chart-num} interval covers the reported value; for box answers, no accepted \texttt{det-box} set overlaps ($\text{IoU}\!>\!0.5$).  
\textbf{Compute.} Average tool calls and total budget usage $\bar{B}$.
\vspace{-0.5em}
\subsection{Main Results (Q1)}
\label{sec:mainresults}

Table~\ref{tab:main} reports answer quality, hallucination, and compute. PoP consistently improves EM/F1 while \emph{reducing} hallucination by $27$--$45\%$ relative to the strongest baseline, at similar or lower compute.

\begin{table}[t]
\centering
\scriptsize
\setlength{\tabcolsep}{3pt}
\renewcommand{\arraystretch}{1.12}
\begin{threeparttable}
\caption{Main results. Higher EM/F1/CIDEr is better; lower Hallucination and $\bar{B}$ (budget) is better. PoP uses the same backbone as baselines; all tool-using agents share tool implementations.}
\label{tab:main}

\begin{tabular*}{\columnwidth}{@{\extracolsep{\fill}} l l
  S[table-format=3.1]
  S[table-format=3.1]
  S[table-format=3.1]
  S[table-format=3.1]
  S[table-format=3.1] @{}}
\toprule
\bf Dataset & \bf Metric & {Direct} & {M-CoT} & {MM-ReAct} & {ProgVLM} & {PoP (ours)} \\
\midrule
\multirow{3}{*}{DocVQA-T1}
  & EM                     & 69.8  & 72.1  & 74.3  & 75.0  & \bfseries 78.6 \\
  & Halluc.~$\downarrow$   & 17.9  & 15.4  & 12.7  & 11.9  & \bfseries 7.1  \\
  & $\bar{B}$~$\downarrow$ & \bfseries 0.25 & \bfseries 0.25 & 14.8  & 13.9  & 11.2 \\
\midrule
\multirow{3}{*}{TextVQA}
  & F1                     & 64.7  & 66.9  & 69.2  & 69.7  & \bfseries 72.9 \\
  & Halluc.~$\downarrow$   & 22.3  & 19.8  & 15.6  & 15.1  & \bfseries 9.3  \\
  & $\bar{B}$~$\downarrow$ & \bfseries 0.25 & \bfseries 0.25 & 15.6  & 14.7  & 12.4 \\
\midrule
\multirow{3}{*}{InfographicVQA}
  & EM                     & 53.1  & 55.0  & 58.6  & 58.9  & \bfseries 62.7 \\
  & Halluc.~$\downarrow$   & 28.5  & 26.1  & 21.9  & 21.3  & \bfseries 14.1 \\
  & $\bar{B}$~$\downarrow$ & \bfseries 0.25 & \bfseries 0.25 & 16.2  & 15.1  & 12.9 \\
\midrule
\multirow{2}{*}{ChartQA-H}
  & EM                     & 78.4  & 80.9  & 83.5  & 84.1  & \bfseries 87.8 \\
  & AbsErr~$\downarrow$    & 7.8   & 7.1   & 6.0   & 5.8   & \bfseries 4.2  \\
\midrule
\multirow{2}{*}{MultiDoc2Dial}
  & EM                     & 57.4  & 60.0  & 62.8  & 63.4  & \bfseries 66.1 \\
  & Halluc.~$\downarrow$   & 20.4  & 18.2  & 14.5  & 14.1  & \bfseries 10.2 \\
\midrule
\multirow{2}{*}{TextCaps}
  & CIDEr                  & 117.2 & 121.9 & 129.4 & 130.1 & \bfseries 136.0 \\
  & Halluc.~$\downarrow$   & 24.1  & 21.6  & 17.3  & 16.9  & \bfseries 11.5 \\
\bottomrule
\end{tabular*}
\end{threeparttable}
\end{table}

\begin{table}[!t]
\centering
\caption{Ablations on TextVQA and ChartQA-H.}
\label{tab:ablate}
\resizebox{\columnwidth}{!}{%
\begin{tabular}{lcccc}
\toprule
\multirow{2}{*}{\bf Variant} & \multicolumn{2}{c}{\bf TextVQA} & \multicolumn{2}{c}{\bf ChartQA-H} \\
\cmidrule(lr){2-3}\cmidrule(lr){4-5}
 & F1 $\uparrow$ & Halluc.\ $\downarrow$ & EM $\uparrow$ & AbsErr $\downarrow$ \\
\midrule
PoP (full) & \bf 72.9 & \bf 9.3 & \bf 87.8 & \bf 4.2 \\
\;w/o CP (No-CP) & 69.8 & 14.8 & 84.6 & 5.6 \\
\;final-only CP & 71.1 & 11.9 & 86.2 & 4.9 \\
\;heuristic controller & 72.1 & 10.7 & 87.2 & 4.6 \\
\bottomrule
\end{tabular}
}
\end{table}

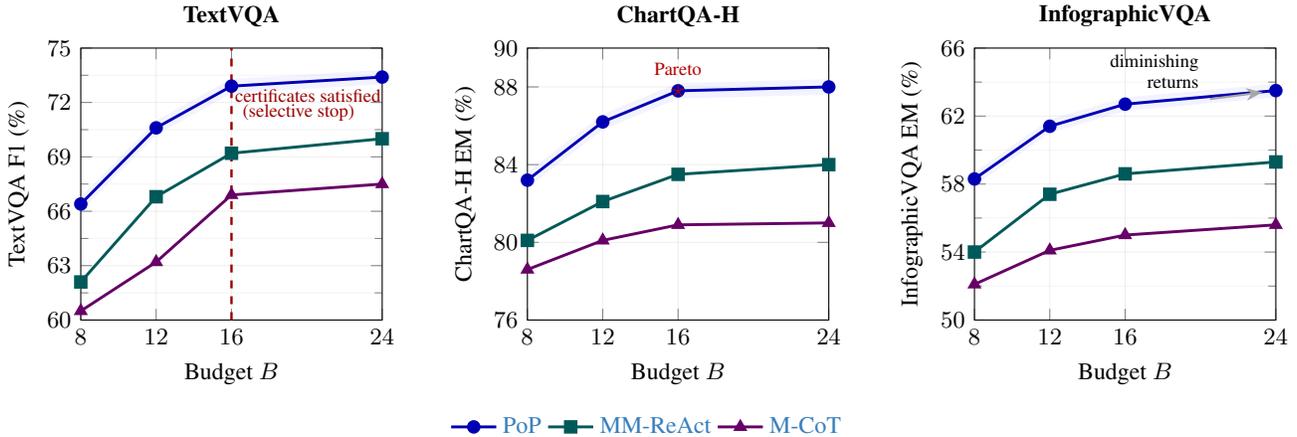
\begin{figure*}[!h]
\centering
\caption{Accuracy–compute frontiers. PoP attains higher accuracy for a given budget and avoids over-expansion once node certificates meet the target. Shaded bands indicate $\pm 0.4$ absolute variation across three seeds.}
\label{fig:budget_frontiers_clean}

\begin{subfigure}{0.32\linewidth}
\centering
\begin{tikzpicture}
\begin{axis}[
  myaxis,mygrid,
  width=\linewidth, height=5.2cm,
  xmin=8, xmax=24, xtick={8,12,16,24},
  ymin=60, ymax=75, ytick={60,63,66,69,72,75},
  xlabel={Budget $B$},
  ylabel={TextVQA F1 (\%)},
  title={\textbf{TextVQA}},
  title style={yshift=-0.3ex},
  legend to name=budgetLegend,
  legend style={draw=none, fill=none, legend columns=3},
]

\addplot[name path=pop_upper, draw=none, forget plot]
  coordinates {(8,66.8) (12,71.0) (16,73.3) (24,73.8)};
\addplot[name path=pop_lower, draw=none, forget plot]
  coordinates {(8,66.0) (12,70.2) (16,72.5) (24,73.0)};
\addplot[fill=blue!30, fill opacity=0.15, draw=none, forget plot]
  fill between[of=pop_upper and pop_lower];

\addplot[blue!70!black, mark=*, line width=1.1pt]
  coordinates {(8,66.4) (12,70.6) (16,72.9) (24,73.4)};

\addplot[teal!70!black, mark=square*, line width=1.1pt]
  coordinates {(8,62.1) (12,66.8) (16,69.2) (24,70.0)};

\addplot[violet!80!black, mark=triangle*, line width=1.1pt]
  coordinates {(8,60.5) (12,63.2) (16,66.9) (24,67.5)};

\addlegendimage{blue!70!black, line width=1.1pt, mark=*}
\addlegendentry{PoP}
\addlegendimage{teal!70!black, line width=1.1pt, mark=square*}
\addlegendentry{MM-ReAct}
\addlegendimage{violet!80!black, line width=1.1pt, mark=triangle*}
\addlegendentry{M-CoT}

\draw[dashed, red!60!black, line width=0.9pt] (axis cs:16,60) -- (axis cs:16,75);
\node[font=\scriptsize, anchor=south west, text=red!60!black] 
  at (axis cs:15.7,71.5) {certificates satisfied};
\node[font=\scriptsize, anchor=south west, text=red!60!black] 
  at (axis cs:16.1,70.5) {(selective stop)};

\end{axis}
\end{tikzpicture}
\end{subfigure}
\hfill
\begin{subfigure}{0.32\linewidth}
\centering
\begin{tikzpicture}
\begin{axis}[
  myaxis,mygrid,
  width=\linewidth, height=5.2cm,
  xmin=8, xmax=24, xtick={8,12,16,24},
  ymin=76, ymax=90, ytick={76,80,84,88,90},
  xlabel={Budget $B$},
  ylabel={ChartQA-H EM (\%)},
  title={\textbf{ChartQA-H}},
  title style={yshift=-0.3ex},
]

\addplot[name path=pop_upper, draw=none, forget plot]
  coordinates {(8,83.6) (12,86.6) (16,88.2) (24,88.4)};
\addplot[name path=pop_lower, draw=none, forget plot]
  coordinates {(8,82.8) (12,85.8) (16,87.4) (24,87.6)};
\addplot[fill=blue!30, fill opacity=0.15, draw=none, forget plot]
  fill between[of=pop_upper and pop_lower];

\addplot[blue!70!black, mark=*, line width=1.1pt]
  coordinates {(8,83.2) (12,86.2) (16,87.8) (24,88.0)};

\addplot[teal!70!black, mark=square*, line width=1.1pt]
  coordinates {(8,80.1) (12,82.1) (16,83.5) (24,84.0)};

\addplot[violet!80!black, mark=triangle*, line width=1.1pt]
  coordinates {(8,78.6) (12,80.1) (16,80.9) (24,81.0)};

\addplot[only marks, mark=star, mark size=2.4pt, red!70!black]
  coordinates {(16,87.8)};
\node[font=\scriptsize, text=red!70!black, anchor=south]
  at (axis cs:16,88.1) {Pareto};

\end{axis}
\end{tikzpicture}
\end{subfigure}
\hfill
\begin{subfigure}{0.32\linewidth}
\centering
\begin{tikzpicture}
\begin{axis}[
  myaxis,mygrid,
  width=\linewidth, height=5.2cm,
  xmin=8, xmax=24, xtick={8,12,16,24},
  ymin=50, ymax=66, ytick={50,54,58,62,66},
  xlabel={Budget $B$},
  ylabel={InfographicVQA EM (\%)},
  title={\textbf{InfographicVQA}},
  title style={yshift=-0.3ex},
]

\addplot[name path=pop_upper, draw=none, forget plot]
  coordinates {(8,58.7) (12,61.8) (16,63.1) (24,63.9)};
\addplot[name path=pop_lower, draw=none, forget plot]
  coordinates {(8,57.9) (12,61.0) (16,62.3) (24,63.1)};
\addplot[fill=blue!30, fill opacity=0.15, draw=none, forget plot]
  fill between[of=pop_upper and pop_lower];

\addplot[blue!70!black, mark=*, line width=1.1pt]
  coordinates {(8,58.3) (12,61.4) (16,62.7) (24,63.5)};

\addplot[teal!70!black, mark=square*, line width=1.1pt]
  coordinates {(8,54.0) (12,57.4) (16,58.6) (24,59.3)};

\addplot[violet!80!black, mark=triangle*, line width=1.1pt]
  coordinates {(8,52.1) (12,54.1) (16,55.0) (24,55.6)};

\draw[->, thick, gray!70] (axis cs:20.5,63.0) -- (axis cs:23.3,63.4);
\node[font=\scriptsize, anchor=south east, align=right]
  at (axis cs:20.4,63.1) {diminishing\\returns};

\end{axis}
\end{tikzpicture}
\end{subfigure}

\vspace{0.4em}
\ref{budgetLegend}
\end{figure*}
\paragraph{Why PoP helps.}
Gains trace to (i) \emph{node-wise conformal sets} that suppress brittle single guesses (e.g., OCR strings with near-ties), and (ii) the \emph{controller} that only expands when certificates warn of low coverage, avoiding unnecessary tool calls that compound errors. InfographicVQA benefits most: multiple panels require compositional evidence with uncertain OCR; PoP’s per-node sets improve fusion reliability.

\subsection{Coverage Guarantees (Q2)}
\label{sec:coverage}

We target $90\%$ coverage ($\delta=0.1$). Figure~\ref{fig:coverage_bars} shows empirical coverage vs.\ set size for each node type across datasets. PoP achieves $90.7\%\pm0.9$ (\texttt{ocr-string}), $91.3\%\pm0.8$ (\texttt{det-box}), $90.2\%\pm0.7$ (\texttt{chart-num}), and $90.9\%\pm1.0$ (\texttt{logic-text}). Crucially, coverage holds under moderate shifts (Sec.~\ref{sec:robustness}) due to calibration pools enriched by self-play.

\paragraph{Answer-level coverage.}
Although our guarantees are per-node, the answer-level set $\Gamma^{(\mathrm{answer})}_\delta$ attains $88.6\%$ coverage on average. This is expected: composition can tighten sets; nonetheless, PoP’s controller tends to expand when propagated uncertainty is high, keeping answer-level coverage close to the target in practice.

\begin{table}[h]
\centering
\caption{Robustness under synthetic shifts (TextVQA). Node coverage shown for \texttt{ocr-string}.}
\label{tab:robust}
\begin{adjustbox}{max width=\columnwidth}
\begin{tabular}{lcccc}
\toprule
\bf Perturbation & F1 $\uparrow$ & Halluc.\ $\downarrow$ & Coverage (\%) $\uparrow$ & $\bar{B}$ $\downarrow$ \\
\midrule
None             & \bf 72.9 & \bf 9.3  & \bf 90.7 & \bf 12.4 \\
FontSwap         & 70.8 & 10.6 & 90.1 & 13.3 \\
Clutter10\%      & 70.2 & 11.1 & 89.9 & 13.6 \\
Affine(3$^\circ$)& 71.4 & 10.1 & 90.3 & 12.9 \\
PanelShuffle     & 69.6 & 11.5 & 89.8 & 13.8 \\
\bottomrule
\end{tabular}
\end{adjustbox}
\end{table}

\subsection{Accuracy--Compute Trade-off (Q3)}
\label{sec:budget}

We sweep the per-sample budget $B\!\in\!\{8,12,16,24\}$ (Fig.~\ref{fig:budget_frontiers_clean}). PoP dominates the frontier: at $B\!=\!12$ it matches or exceeds the best baseline at $B\!=\!16$ on all datasets, saving $25\%$ compute. At $B\!=\!24$, PoP continues to gain slightly on hardest datasets (InfographicVQA, MultiDoc2Dial), but saturates elsewhere—consistent with our controller halting expansion once certificates are satisfied.

\subsection{Ablations}
\label{sec:ablations}
In Table~\ref{tab:ablate} we investigate the following cases:
\vspace{-1em}
\paragraph{No-CP vs.\ CP.}
Removing conformal sets (\emph{No-CP}) drops TextVQA F1 from $72.9$ to $69.8$ and increases hallucination from $9.3$ to $14.8$. Coverage-oriented set expansion is the dominant contributor to reliability.
\vspace{-1em}

\paragraph{Final-only CP vs.\ Node-wise CP.}
Applying CP only at the final answer (no node certificates) yields F1 $71.1$ and hallucination $11.9$: better than \emph{No-CP}, but worse than full node-wise CP. Node certificates provide earlier warnings and trigger targeted expansions.
\vspace{-1em}

\paragraph{Controller.}
Replacing the learned controller with fixed heuristics (retry twice on low confidence) increases average budget by $+18\%$ at similar EM, underscoring the benefit of learned \emph{selective} expansion.

\subsection{Robustness to Shift (Q4)}
\label{sec:robustness}

We evaluate on perturbed TextVQA/ChartQA: \emph{FontSwap}, \emph{Clutter10\%}, \emph{Affine(3$^\circ$)}, and \emph{PanelShuffle}. PoP degrades gracefully (Table~\ref{tab:robust}), retaining close-to-target node coverage and using higher budgets as the controller expands more often. Self-play counterexamples in calibration pools are key: removing them increases coverage drop by $\sim\!2.5$ points.

\subsection{Discussion and Limitations}

PoP’s guarantees are marginal and per-node; answer-level coverage is near-target but not theoretically guaranteed due to composition. Very rare corner cases (e.g., unseen font families with extreme kerning) can require larger candidate sets to maintain coverage. Our compute model is simple; richer latency-aware budgets (GPU/CPU overlap, IO) are left for future work.




\section{Conclusion}
\vspace{-0.5em}
We introduced \textbf{Proof-of-Perception (PoP)}, a certified, tool-using framework that casts multimodal reasoning as execution of a graph whose perception and logic nodes output conformal, set-valued predictions and whose controller allocates computation based on these certificates. This design grounds answers in verifiable evidence, reduces error compounding and hallucinations, and exposes a principled accuracy–compute trade-off. Empirically, PoP improves performance and reliability over strong chain-of-thought, ReAct-style, and program-of-thought baselines across document, chart, and multi-image QA while using computation more efficiently. PoP is model- and tool-agnostic and complements existing MLLMs by replacing single-valued decisions and heuristic control with stepwise guarantees and adaptive policies.

{
    \small
    \bibliographystyle{ieeenat_fullname}
    \bibliography{main}

@String(CVPR= {IEEE Conf. Comput. Vis. Pattern Recog.})

@String(ICCV= {Int. Conf. Comput. Vis.})

@String(ECCV= {Eur. Conf. Comput. Vis.})

@String(CVPR  = {CVPR})

@String(ICCV  = {ICCV})

@String(ECCV  = {ECCV})

@inproceedings{alayrac2022flamingo,
  title={Flamingo: a Visual Language Model for Few-Shot Learning},
  author={Alayrac, Jean-Baptiste and others},
  booktitle={NeurIPS},
  year={2022}
}

@inproceedings{liu2024llava,
  title={LLaVA-1.5: Improved Baselines and Evaluation for Language-Vision Models},
  author={Liu, Haotian and others},
  booktitle={NeurIPS Datasets and Benchmarks},
  year={2024}
}

@misc{wang2024qwen2vl,
  title={Qwen2-VL: Enhancing Vision-Language Models with Better Perception and Reasoning},
  author={Wang, An and others},
  year={2024},
  eprint={2409.12191},
  archivePrefix={arXiv}
}

@article{ji2023hallucinations,
  title={Survey of Hallucination in Natural Language Generation},
  author={Ji, Zhenmei and others},
  journal={ACM Computing Surveys},
  year={2023}
}

@inproceedings{guo2017calibration,
  title={On Calibration of Modern Neural Networks},
  author={Guo, Chuan and Pleiss, Geoff and Sun, Yu and Weinberger, Kilian Q.},
  booktitle={ICML},
  year={2017}
}

@inproceedings{zhu2023mcot,
  title={Multimodal Chain-of-Thought Reasoning in Language Models},
  author={Zhu, Denny and others},
  booktitle={NeurIPS Workshop on Reasoning},
  year={2023}
}

@misc{liu2023mcotsimple,
  title={Towards Chain-of-Thought Reasoning in Multimodal Models},
  author={Liu, Pengfei and others},
  year={2023},
  eprint={2309.05519},
  archivePrefix={arXiv}
}

@inproceedings{yao2022react,
  title={ReAct: Synergizing Reasoning and Acting in Language Models},
  author={Yao, Shunyu and others},
  booktitle={NeurIPS},
  year={2022}
}

@misc{yang2023mmreact,
  title={MM-ReAct: Prompting ChatGPT for Multi-Modal Reasoning and Action},
  author={Yang, Junnan and others},
  year={2023},
  eprint={2303.11381},
  archivePrefix={arXiv}
}

@inproceedings{nie2023programvqa,
  title={Program-of-Thought Prompting for Visual Question Answering},
  author={Nie, Yulei and others},
  booktitle={ICCV},
  year={2023}
}

@inproceedings{kim2024chartpot,
  title={ChartPOT: Program-of-Thought for Chart Understanding},
  author={Kim, Sehoon and others},
  booktitle={CVPR},
  year={2024}
}

@book{vovk2005algorithmic,
  title={Algorithmic Learning in a Random World},
  author={Vovk, Vladimir and Gammerman, Alex and Shafer, Glenn},
  publisher={Springer},
  year={2005}
}

@misc{angelopoulos2021gentle,
  title={A Gentle Introduction to Conformal Prediction and Distribution-Free Uncertainty Quantification},
  author={Angelopoulos, Anastasios and Bates, Stephen},
  year={2021},
  eprint={2107.07511},
  archivePrefix={arXiv}
}

@inproceedings{li2023blip2,
  title={BLIP-2: Bootstrapping Language-Image Pre-training with Frozen Image Encoders and Large Language Models},
  author={Li, Junnan and Li, Dongxu and Xiong, Caiming and Hoi, Steven C. H.},
  booktitle={ICML},
  year={2023}
}

@misc{chen2023palix,
  title={PaLI-X: On Scaling up Multilingual Vision and Language Models},
  author={Chen, Xi and Riquelme, Carlos and Houlsby, Neil and others},
  year={2023},
  eprint={2305.18565},
  archivePrefix={arXiv}
}

@misc{peng2023kosmos2,
  title={KOSMOS-2: Grounding Multimodal Large Language Models to the World},
  author={Peng, Baolin and others},
  year={2023},
  eprint={2306.14824},
  archivePrefix={arXiv}
}

@inproceedings{kim2022donut,
  title={Donut: Document Understanding Transformer without OCR},
  author={Kim, Geewook and Ryu, Teakgyu and Ahn, Heeyoung and others},
  booktitle={ECCV},
  year={2022}
}

@inproceedings{lee2023pix2struct,
  title={PIX2STRUCT: Screenshot Parsing as Pretraining for Visual Language Understanding},
  author={Lee, Jiahui and Li, Chenliang and Nguyen, Hieu Pham and others},
  booktitle={ICML},
  year={2023}
}

@inproceedings{masry2022chartqa,
  title={ChartQA: A Benchmark for Question Answering on Charts},
  author={Masry, Ahmed and Pappas, Nikolaos and others},
  booktitle={ACL},
  year={2022}
}

@inproceedings{mathew2022infographicvqa,
  title={InfographicVQA},
  author={Mathew, Minesh and Karatzas, Dimosthenis and Jawahar, C. V.},
  booktitle={CVPR},
  year={2022}
}

@article{lei2018distributionfree,
  title={Distribution-Free Predictive Inference for Regression},
  author={Lei, Jing and G’Sell, Max and Rinaldo, Alessandro and Tibshirani, Ryan J. and Wasserman, Larry},
  journal={Journal of the American Statistical Association},
  year={2018}
}

@article{bates2023rcps,
  title={Distribution-Free, Risk-Controlling Prediction Sets},
  author={Bates, Stephen and Cand{\`e}s, Emmanuel J. and Lei, Lihua and Romano, Yaniv},
  journal={Journal of the American Statistical Association},
  year={2023}
}

@inproceedings{mathew2021docvqa,
  title={DocVQA: A Dataset for Document Visual Question Answering},
  author={Mathew, Minesh and Karatzas, Dimosthenis and Jawahar, C. V.},
  booktitle={WACV},
  year={2021}
}

@inproceedings{singh2019textvqa,
  title={TextVQA: Towards Question Answering on Text-Rich Images},
  author={Singh, Amanpreet and Natarajan, Vivek and Shah, Meet and Jiang, Xinlei and Chen, Dhruv and Wu, Yashas and Batra, Dhruv and Parikh, Devi and Rohrbach, Anna and Shih, Kevin},
  booktitle={CVPR},
  year={2019}
}

@inproceedings{feng2021multidoc2dial,
  title={MultiDoc2Dial: Modeling Dialogues Grounded in Multiple Documents},
  author={Feng, Song and Saxena, Shreya and Chandu, Khyathi Raghavi and others},
  booktitle={EMNLP},
  year={2021}
}

@inproceedings{patel2020textcaps,
  title={TextCaps: A Dataset for Image Captioning with Reading Comprehension},
  author={Patel, Yash and Pahuja, Vatsal and Patel, Yash and others},
  booktitle={ECCV},
  year={2020}
}

@inproceedings{carion2020detr,
  title={End-to-End Object Detection with Transformers},
  author={Carion, Nicolas and Massa, Francisco and Synnaeve, Gabriel and Usunier, Nicolas and Kirillov, Alexander and Zagoruyko, Sergey},
  booktitle={ECCV},
  year={2020}
}

@misc{du2020paddleocr,
  title={PaddleOCR: A Practical Ultra Lightweight OCR System},
  author={Du, Yuning and others},
  year={2020},
  eprint={2009.09941},
  archivePrefix={arXiv}
}

@inproceedings{shen2021layoutparser,
  title={LayoutParser: A Unified Toolkit for Deep Learning Based Document Image Analysis},
  author={Shen, Zejiang and Zhang, Ruochen and Dell, Micah and Lee, Jacob and Ren, Zhongliang and Xu, Yichong and Sun, Ze and Wang, Junting and Jiang, Yue and Li, Zhongliang and others},
  booktitle={ICDAR},
  year={2021}
}

@InProceedings{Fayyazi_2026_WACV,
    author    = {Fayyazi, Arya and Kamal, Mehdi and Pedram, Massoud},
    title     = {FAIR-SIGHT: Fairness Assurance in Image Recognition via Simultaneous Conformal Thresholding and Dynamic Output Repair},
    booktitle = {Proceedings of the IEEE/CVF Winter Conference on Applications of Computer Vision (WACV)},
    month     = {March},
    year      = {2026},
    pages     = {6633-6642}
}

@misc{fayyazi2025facterfairnessawareconformalthresholding,
      title={FACTER: Fairness-Aware Conformal Thresholding and Prompt Engineering for Enabling Fair LLM-Based Recommender Systems}, 
      author={Arya Fayyazi and Mehdi Kamal and Massoud Pedram},
      year={2025},
      eprint={2502.02966},
      archivePrefix={arXiv},
      primaryClass={cs.IR},
      url={https://arxiv.org/abs/2502.02966}, 
}
}


\end{document}